\documentclass[runningheads]{llncs}

\usepackage[final,year=2024,ID=10615]{eccv}
\usepackage{eccvabbrv}

\usepackage{graphicx}
\usepackage{booktabs}
\usepackage{algorithm}
\usepackage{algorithmic}
\usepackage{multirow}
\usepackage{colortbl}
\usepackage{xcolor}
\usepackage{array}
\usepackage[accsupp]{axessibility} % Improves PDF readability for those with disabilities.

\usepackage[pagebackref,breaklinks,colorlinks,citecolor=eccvblue]{hyperref}
\usepackage{hyperref}

\usepackage{orcidlink}

\definecolor{lightcyan}{HTML}{E1FFFF}

\begin{document}

\title{
\raisebox{-0.2cm}{\includegraphics[scale=0.05]{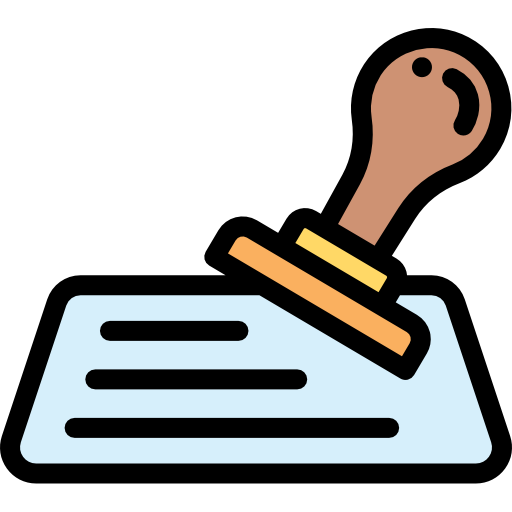}}
STAMP: Outlier-Aware Test-Time Adaptation with Stable Memory Replay}

% TODO REVIEW: If the paper title is too long for the running head, you can set an abbreviated paper title here. If not, comment out.
\titlerunning{Outlier-Aware Test-Time Adaptation with Stable Memory Replay}

\author{Yongcan Yu\inst{1,2}\orcidlink{0009-0000-8261-1662} \and
Lijun Sheng\inst{1,3}\orcidlink{0000-0002-8240-9736} \and
Ran He\inst{1,2}\orcidlink{0000−0002−3807−991X} \and
Jian Liang\inst{1,2}\thanks{Corresponding author.}\orcidlink{0000−0003−3890−1894}}

\authorrunning{Yu et al.}

\institute{NLPR \& MAIS, Institute of Automation, Chinese Academy of Sciences \\ 
\and School of Artificial Intelligence, University of Chinese Academy of Sciences 
\and University of Science and Technology of China \\
\email{\{yuyongcan0223, liangjian92\}@gmail.com}
}
\maketitle

\begin{abstract}
Test-time adaptation (TTA) aims to address the distribution shift between the training and test data with only unlabeled data at test time.
Existing TTA methods often focus on improving recognition performance specifically for test data associated with classes in the training set.
However, during the open-world inference process, there are inevitably test data instances from unknown classes, commonly referred to as outliers.
This paper pays attention to the problem that conducts both sample recognition and outlier rejection during inference while outliers exist.
To address this problem, we propose a new approach called STAble Memory rePlay (STAMP), which performs optimization over a stable memory bank instead of the risky mini-batch.
In particular, the memory bank is dynamically updated by selecting low-entropy and label-consistent samples in a class-balanced manner.
In addition, we develop a self-weighted entropy minimization strategy that assigns higher weight to low-entropy samples.
Extensive results demonstrate that STAMP outperforms existing TTA methods in terms of both recognition and outlier detection performance.
The code is released at \url{https://github.com/yuyongcan/STAMP}.
\end{abstract}

\section{Introduction}
\label{intro}
\begin{figure}[!t]
\begin{center}
\centerline{\includegraphics[width=\columnwidth]{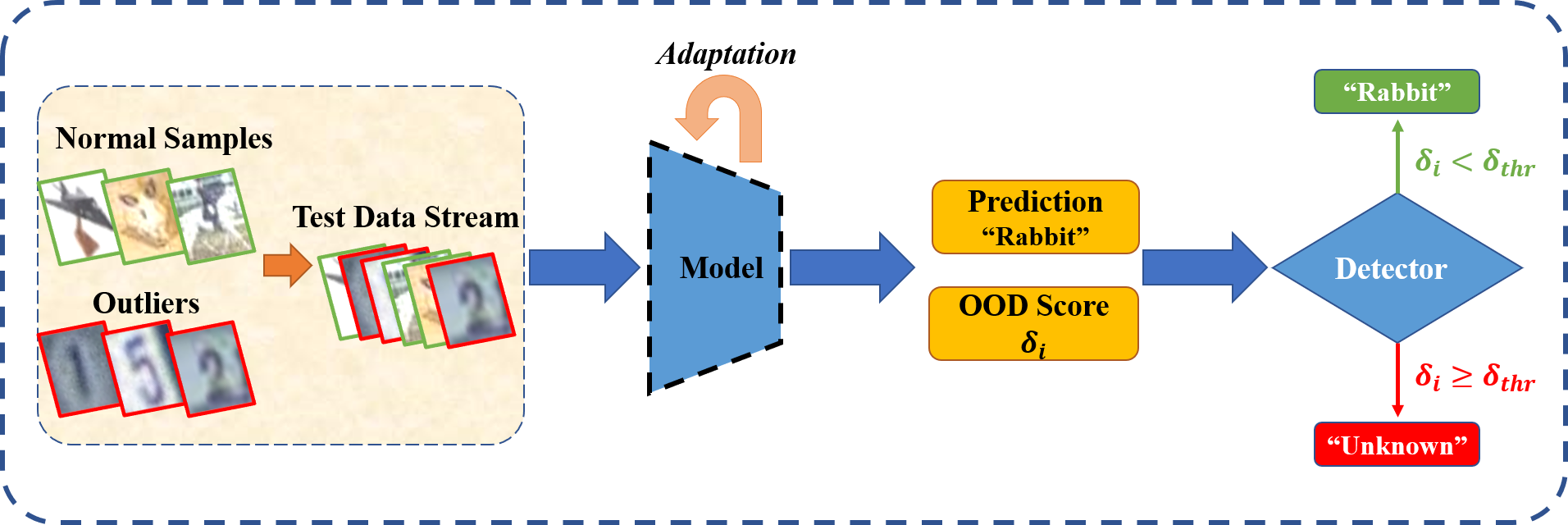}}
\caption{The illustration of outlier-aware test-time adaptation. The test data stream consists of both normal samples and outliers. When the test data arrives, the algorithm adapts the model and provides two outputs, a prediction and an OOD score. OOD score indicates the likelihood that a sample is an outlier. The detector in the deployment scenario chooses to keep the prediction or reject samples according to the OOD score.}
\label{fig:framework}
\end{center}
\end{figure}

Deep neural networks (DNNs) \cite{tan2019efficientnet,vaswani2017attention} have achieved remarkable performance across various tasks.
Nevertheless, the impressive performance is based on the assumption that training data and test data come from the same distribution.
However, in real-world scenarios, the presence of distribution shifts is ubiquitous \cite{farahani2021brief, liang2020we}, leading to a degradation in the prediction performance \cite{croce2021robustbench}.
Taking autonomous driving \cite{yurtsever2020survey} as an example, changing weather conditions causes systems trained on laboratory data to make incorrect decisions.
To mitigate the impact of distribution shift, test-time adaptation (TTA) \cite{liang2023comprehensive,yu2023benchmarking} has emerged as a promising approach, adapting models with unlabeled samples during inference.
Nowadays, TTA has achieved remarkable results in various tasks, such as image classification \cite{wang2020tent, zhang2023adanpc}, semantic segmentation \cite{cao2023multi, volpi2022road}, object detection \cite{an2023context}.

Previous TTA works \cite{niu2022efficient, yuan2023robust, chen2022contrastive} always assume that test data and training data share the same class space.
Nevertheless, in realistic open-world scenarios, test data streams will inevitably contain samples from new semantic classes, also known as outliers.
The introduction of outliers brings two additional risks to adaptation tasks:
(i) Classifying outliers into known categories poses safety risks, especially in contexts like autonomous driving, where systems should request human intervention to handle unknown objects to avoid traffic accidents.
(ii) Self-supervised loss (\eg, entropy minimization, pseudo-labeling) calculated with outliers may misguide the optimization process, decreasing the model's recognition ability on known classes.

To better tackle the above risks, we pay attention to a more realistic TTA scenario termed outlier-aware test-time adaptation.
Outlier-aware test-time adaptation aims to conduct recognition for normal samples and reject to make predictions for outliers.
The overview of outlier-aware test-time adaptation is illustrated in \Cref{fig:framework}.
When test data mixed with outliers arrives, the algorithm adapts the model and generates the prediction and an out-of-distribution (OOD) score for each sample.
Then we decide to keep the prediction or reject the sample based on its OOD score.
The sample with a higher OOD score is more likely to be considered an outlier.

To address the outlier-aware test-time adaptation, we additionally design a novel algorithm called STAble Memory rePlay (STAMP).
STAMP consists of three components: reliable class-balanced memory, self-weighted entropy minimization, and stable optimization strategy.
Reliable class-balanced memory stores reliable samples for optimization by filtering samples based on their prediction's consistency and entropy.
To maintain the class-balanced prediction of the model, it dynamically discards samples with the class that participates most frequently in optimization when the memory is full.
Furthermore, we adopt self-weighted entropy minimization to assign greater weights to low-entropy samples in memory and adjust the weights for each sample adaptively.
To ensure the stability of the optimization process, we dynamically adjust the optimization step size to quickly adapt to the target domain at the early stage and mitigate error accumulation later.
Moreover, we apply the sharpness-aware minimization technique\cite{foret2020sharpness} to further reduce the impact of noisy gradients.

To demonstrate the effectiveness of STAMP, we compare it with eight state-of-the-art TTA algorithms. We select three commonly used datasets in TTA, including CIFAR10-C, CIFAR100-C \cite{devries2018learning,hendrycks2019benchmarking}, and ImageNet-C \cite{deng2009imagenet,hendrycks2019benchmarking}, alongside some outlier datasets (i.e., Noise, SVHN \cite{netzer2011reading}, LSUN \cite{yu2015lsun}, TinyImageNet \cite{le2015tiny}, Textures \cite{cimpoi2014describing}, Places365 \cite{zhou2017places}).
Experimental results show that our method achieves superior performance on recognition and outlier detection under the outlier-aware TTA setting.
The experiment on conventional protocol indicates that STAMP also performs well in the scenario without outliers.
In summary, we list our contributions as follows:
\begin{itemize}
  \item We introduce outlier-aware test-time adaptation, a TTA scenario considering both normal sample recognition and outlier rejection during inference.
  \item We propose a memory that dynamically stores reliable samples for replay. To prevent noisy samples from entering the memory, we propose a filtering strategy based on the consistency and entropy of the model's prediction.
  \item We design self-weighted entropy minimization, which guides the model to adaptively learn to assign higher weights to valuable samples.
  \item We propose a proper metric that considers both generalization and outlier detection performance. Extensive experiments compared with eight TTA algorithms on several benchmarks demonstrate the effectiveness of our method.
\end{itemize}

\section{Related Work}
\subsection{Test-Time Adaptation}
TTA intends to adapt the model pre-trained on the source domain to the target domain while the distribution shift exists.
Compared to domain generalization \cite{wang2022generalizing, tong2023distribution} and domain adaptation \cite{gebru2017fine, wang2018deep}, TTA needs to access neither the source data nor the target labels, making it has a wide range of applications \cite{ding2023maps, ding2023proxymix, sheng2023adaptguard}.
TTA methods can be mainly divided into two solutions: batch normalization (BN) calibration and self-training.
In BN calibration, Nado et al.\cite{nado2020evaluating} utilize the BN statistics of the test batch ($\mu_t, \sigma_t$) to conduct normalization in the BN layer,
while Schneider \cite{schneider2020improving} propose a more comprehensive approach considering statistics from both the source domain ($\mu_s, \sigma_s$) and the target domain to calibrate the domain shifts.
To conquer the correlation sampling problem, Yuan et al. \cite{yuan2023robust} adopt robust batch normalization, which updates the statistic ($\mu_t, \sigma_t$) in a moving average manner with the test data stream.
Currently, the primary paradigm in TTA is self-training \cite{zhang2022memo, boudiaf2022parameter}, where Wang et al.\cite{wang2020tent} adopt entropy minimization as a self-supervised loss for optimization and Chen et al.\cite{chen2022contrastive} introduce contrastive loss to TTA. However, unsupervised loss may contain noise information.
Niu et al.\cite{niu2022efficient} first propose filtering out high-entropy samples, a mechanism adopted by subsequent works \cite{niu2022towards, gong2023sotta}.
Subsequently, more attention has been paid to addressing more realistic adaptation scenarios such as continually changing target domains \cite{wang2022continual, chakrabarty2023sata}, label shift \cite{zhou2023ods, park2023label}, and temporally correlated shift \cite{zhao2023pitfalls,gong2022note}.
In our study, we focus on a more realistic scenario where test data contains samples with new semantic classes.
Some existing works \cite{lee2023towards, gong2023sotta, li2023robustness} discuss the TTA algorithm in scenarios containing outliers.
However, Lee et al.\cite{lee2023towards} and Gong et al.\cite{gong2023sotta} solely focus on the generalization ability to domain shift, neglecting the issue of rejecting the outliers.
A recent work \cite{li2023robustness} discusses outlier detection performance based on TTA tasks, but it only provides binary predictions to indicate whether the test data is outlier.
In contrast, outlier-aware test-time adaptation is more flexible, providing an OOD score, and different systems can conduct sample rejection based on security requirements and the score.

\subsection{Out-of-Distribution Detection}
OOD detection aims to identify instances that significantly deviate from the distribution of the training data \cite{liangrealistic}.
Their primary focus is usually on detecting samples whose labels were not seen during the training phase.
This area has been extensively researched, with one common approach being the utilization of confidence scores such as max softmax probability \cite{hendrycks2016baseline}, energy score \cite{liu2020energy, lin2021mood}, and others \cite{lee2018simple} derived from the model's output to detect OOD samples.
Further improvements have been made by Liang et al.\cite{liang2017enhancing} through techniques like temperature scaling and input preprocessing to enhance the effectiveness of confidence scores.
Another avenue for OOD sample detection lies in training-based methods \cite{wang2021energy, devries2018learning, vyas2018out}.
These techniques involve training a branch using both in-distribution data and OOD data to enable the model to distinguish OOD data during training.
However, these methods alter the training process of the source model, which can not be directly applied to adaptation tasks based on a well-trained model.
Moreover, OOD detection traditionally assumes that both the training and test data originate from the same domain, a condition not met in TTA scenarios.
Under the realistic scenario, there may be a domain shift between test data and training data, which brings greater challenges to detection.

\section{Methodology}
In this section, we first introduce the outlier-aware test-time adaptation problem in \Cref{sec:owtta}.
Then we describe our proposed method, including how to build a memory bank and conduct the optimization in \Cref{sec:sem}.

\subsection{Outlier-Aware Test-Time Adaptation}
\label{sec:owtta}
Let $\mathcal{D}_S$ be the source domain dataset with label space $\mathcal{C}_S$ follows a distribution $P_S(x,y)$. Similarly, let $\mathcal{D}_T=\{(x_i, y_i)\}_{i=1}^{N_{T}}$ be the target domain dataset with label space $\mathcal{C}_T$ follows a distribution $P_T(x,y)$.
Given a source model $f_{\theta_s}(x)$ well trained on $\mathcal{D}_S$ with parameter $\theta_s$. TTA aims to adapt the source model to the target domain while $P_S(x) \neq P_T(x)$.

In traditional closed-set TTA, it is commonly assumed that the $\mathcal{D}_S$ and $\mathcal{D}_T$ share the same label space, i.e., $\mathcal{C}_S = \mathcal{C}_T$.
However, in outlier-aware test-time adaptation, we consider $\mathcal{C}_S \subseteq \mathcal{C}_T$ while some outliers with new semantic classes will be encountered in $\mathcal{D}_T$.
The illustration of outlier-aware test-time adaption is shown in \Cref{fig:framework}.
In addition to adapting the model and providing the prediction, we also generate an OOD score $\delta_i$ for each sample $x_i$.
Then we adopt an outlier detector $g(\cdot)$ as follow:
\begin{equation}
	g(x_i) = \begin{cases}
	1, &\text{if}\ \delta_i \ge \delta_{thr}\\
	0, &\text{if}\ \delta_i < \delta_{thr},
		  \end{cases}
\end{equation}
where $\delta_{thr}$ is a threshold that can be adjusted according to the requirements of different application scenarios.
The samples with $g(x_i)=1$ will be considered outliers and rejected for recognition.
And we output the prediction for samples with $g(x_i)=0$.
In our work, we take the entropy of the model prediction as our OOD score.

\begin{figure*}[!t]
\begin{center}
\centerline{\includegraphics[width=\textwidth]{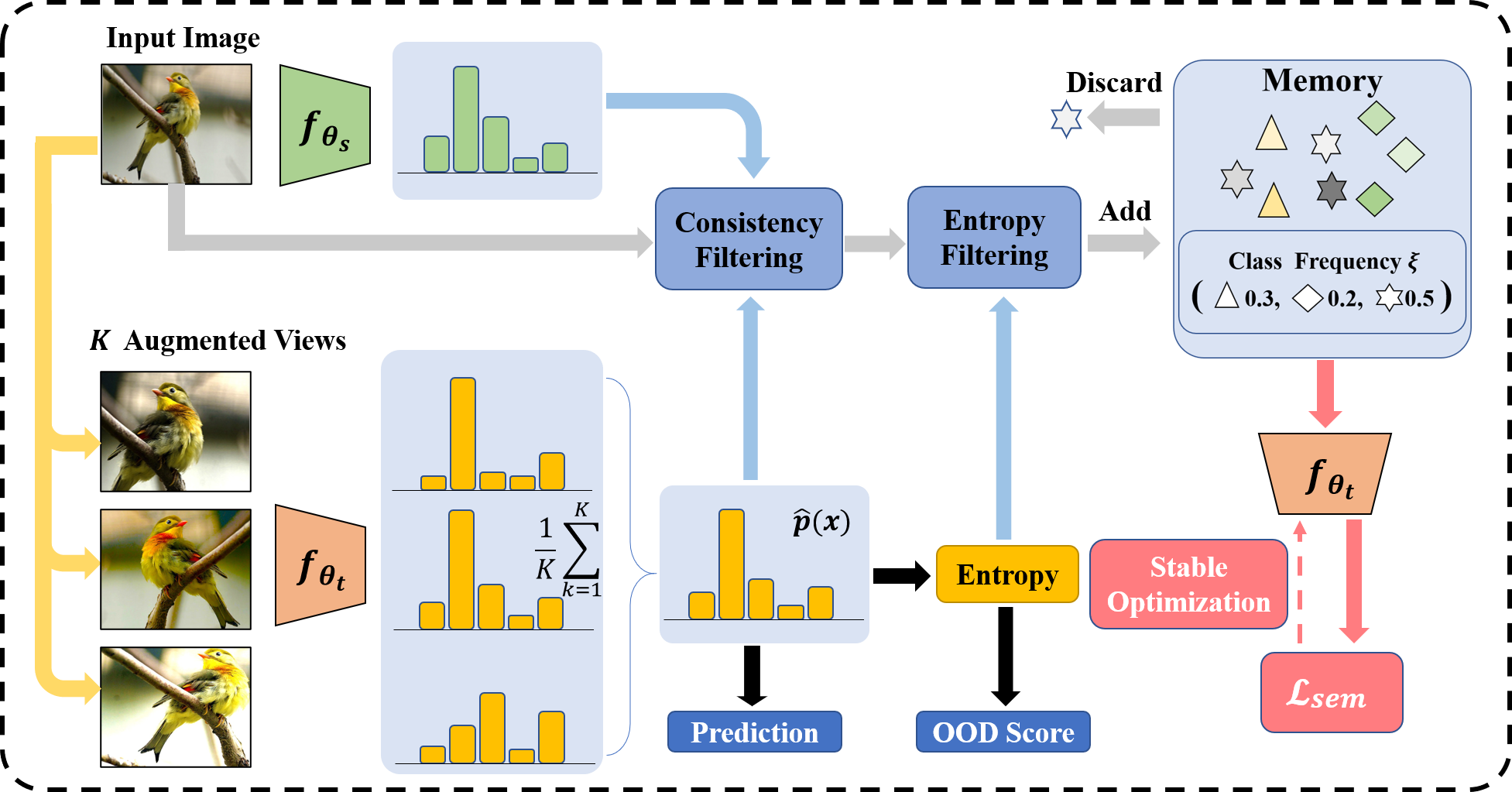}}
\caption{The overview of STAMP. STAMP retains reliable samples for optimizing the network by maintaining a memory bank. When test samples arrive, STAMP generates an augmentation-averaged prediction. Two filtering mechanisms are designed to filter out inconsistent and high-entropy samples and add the remaining to the memory. A frequency vector is maintained to select samples to discard when the memory is full. STAMP leverages the reliable samples in the memory to optimize the model with the SAM optimizer alongside a decaying step strategy.}
\label{fig:overview}
\end{center}
\end{figure*}

\subsection{Stable Memory Replay}
\label{sec:sem}
In this section, we introduce stable memory replay (STAMP). STAMP first uses a filtering mechanism to filter out unreliable samples in the coming mini-batch and stores reliable samples in memory. Then, STAMP optimizes the model using in-memory samples by applying self-weighted entropy minimization. In order to further improve the stability of optimization, STAMP uses test-time step size decay and sharpness-aware minimization to guide optimization. The illustration of STAMP is shown in \Cref{fig:overview}.

\begin{algorithm}[tb]
  \caption{Reliable Class-Balanced Memory}
  \label{alg:RBM}
\begin{algorithmic}
  \STATE {\bfseries Input:} memory bank $\mathcal{M}$, a batch of data $\{x_i\}_{i=1}^B$, entropy threshold $\mathcal{H}_{thr}$, class frequency $\{\xi_c\}_{c=1}^{\lvert\mathbf{C}_S\rvert}$.
  \FOR{$i=1$ {\bfseries to} $B$}
  \STATE Calculate $\hat{p}(x_i)$ by \Cref{equ:pred} and $\hat{y}_i = \arg\max_c [\hat p(x_i)]_c$.
  \STATE Calculate $m_{con}(x_i)$, $m_{ent}(x_i)$ of $x_i$ by Equation \eqref{equ:mcon}, \eqref{equ:ment}.
  \IF{$m_{con}(x_i) = 1$ and $m_{ent}(x_i)=1$}
  \IF{$\mathcal{M}$ is full} 
  \STATE Discard one sample with highest class frequency $\xi$ \ from $\mathcal{M}$.
  \ENDIF
  \STATE add $(x_i, \hat{y}_i)$ into $\mathcal{M}$.
  \ENDIF
  \ENDFOR
  \STATE Update $\xi$ with Equation \eqref{equ:frequency}.
\end{algorithmic}
\end{algorithm}

\textbf{Reliable Class-Balanced Memory.}\quad
Since excluding unreliable samples from optimization has shown great effectiveness \cite{niu2022efficient, gong2023sotta}, we propose reliable class-balanced memory to store reliable samples for optimization.
To have a more robust estimation of the sample's reliability, we get its prediction in an augmentation-averaged way.
Moreover, we prevent unreliable samples from getting into the memory bank $\mathcal{M}$ by employing consistency filtering and entropy filtering.
To preserve the model's balanced class prior distribution, we dynamically discard samples with the highest class frequency from the memory bank.

To ensure robust predictions, we adopt the augmentation-averaged method, which addresses the issue of deep neural networks being overconfident with out-of-distribution samples. This overconfidence often blurs the line between in-distribution and out-of-distribution samples. However, research has demonstrated that employing test-time augmentation \cite{zhang2022memo, shanmugam2021better} enhances model calibration \cite{nixon2019measuring}. Thus, when presented with a test sample $x$, we derive its prediction through the following process:
\begin{equation}
\label{equ:pred}
  % \hat p_t = \frac{1}{N_{aug}}\sum_{i=1}^{N_{aug}} f_{\theta_t}(aug_i(x_t)),
  \hat{p}(x) = \frac{1}{K}\sum_{k=1}^{K} f_{\theta_t}(\tau_k(x)),
\end{equation}
where $\tau(\cdot)$ is the operation of augmentation, $K$ is the total number of augmented views.
By averaging predictions over augmented samples, the model can better eliminate the uncertainty inherent in its predictions, improving calibration and obtaining more reliable confidence estimates.
In \Cref{sec:aug}, we discuss the impact of the augmentation-averaged estimation.

After acquiring the prediction $\hat{p}$, we implement filtering strategies to prevent unreliable samples from being added to memory for two reasons. Firstly, optimization using unreliable samples can result in error accumulation over time \cite{niu2022efficient}. Secondly, we prevent outliers from entering the memory and reduce the entropy of normal samples in the memory. This action facilitates the model's ability to distinguish between normal samples and outliers. 
Since reliable samples that are beneficial to optimization tend to have the same pseudo-label as the source model, we employ consistency filtering to remove samples whose predictions are inconsistent with those of the source model:
\begin{equation}
\label{equ:mcon}
  m_{con}(x) = \mathbb{I} \Big(\arg\max_{c} \ [\hat{p}(x)]_c=\arg\max_{c}\ [f_{\theta_s}(x)]_c \Big),
\end{equation}
where $\mathbb{I}(\cdot)$ is the indicator function.
Moreover, including high entropy samples for optimization can harm the stability of the model \cite{niu2022efficient, niu2022towards}, while samples with low entropy always possess greater value and reliability.
Therefore, we use entropy filtering to filter out samples with high entropy from entering the memory.
\begin{equation}
\label{equ:ment}
  m_{ent}(x) = \mathbb{I} \Big(\mathcal{H}(\hat{p}(x))< \mathcal{H}_{thr} \Big) ,
\end{equation}
where $\mathcal{H}(q)=-\sum_c q_c\log q_c$ denotes the entropy of the prediction vector $q$.
We filter out samples with $\mathcal{H}(\hat p) \ge \mathcal{H}_{thr}$, where $\mathcal{H}_{thr}$ is a threshold.
Considering both filtering strategies, we add sample $(x, \hat{y})$ into memory bank $\mathcal{M}$ if the sample satisfies consistency filtering and entropy filtering simultaneously, \ie, $m_{con}(x) = 1$ and $m_{ent}(x) = 1$.

Since the number of samples stored in memory is limited, we design a class-balanced strategy to choose a sample to discard when the memory is full.
Since we adopt entropy minimization as our optimization objective, an issue arises when the model excessively optimizes samples from a single category over a short period.
It leads to a bias towards predicting the dominant category.
To counteract this bias, we dynamically discard samples from the memory that belong to the category with the highest update frequency.
After each optimization of the model, we update the class frequencies in a moving average manner:
\begin{equation}
\label{equ:frequency}
  \xi_{c}^t = (1-\beta)\xi_{c}^{t-1} + \beta\nu_c^t,
\end{equation}
where $\nu_c^t$ represents the number of samples with class label $c$ in memory and $\beta$ is moving average coefficient hyperparameter.
While the memory is full, we discard a sample from the class with the highest $\xi$.
We summarize the work process of reliable class-balanced memory in \Cref{alg:RBM}.

\begin{algorithm}[tb]
  \caption{The Pipeline of STAMP}
  \label{alg:STAMP}
\begin{algorithmic}
\STATE {\bfseries Initialize:} Memory bank $\mathcal{M} = \emptyset$, zero-initialized class frequency
$\{\xi_c\}_{c=1}^{\lvert \mathcal{C}_S \rvert}$. 
% $\xi_c=0$ for $c\in 1 \text{ to } \lvert \mathcal{C}_S \rvert$.
\STATE {\bfseries Input:} Unlabeled test samples $\mathcal{D}_T = \{x_i\}_{i=1}^{N_{T}}$, a trained model $f_{\theta_s}(\cdot)$.
  \FOR{each online mini-batch $X_b$}
  \STATE Calculate $\hat{p}(x)$ by Equation \eqref{equ:pred} and $\mathcal{H}(\hat{p}(x)$) for all $x\in X_b$.
  \STATE Obtain the prediction $\hat{y}=\arg\max_c [\hat{p}(x)]_c$ for all $x\in X_b$.
  % \STATE $\hat{y} := \arg\max \nolimits_c \hat{p_c}$
  \STATE Update memory bank $\mathcal{M}$ by \Cref{alg:RBM}.
  \STATE Compute $\mathcal{L}_{sem}$ with $\mathcal{M}$ by Equation \eqref{equ:loss}.
  \STATE Update $\theta_t$ with SAM in Equation~\eqref{equ:sam} and decay step size in Equation~\eqref{equ:ss}.
  \ENDFOR
\STATE {\bfseries Output:} The prediction and OOD score $\{\hat{y}_i, \mathcal{H}(\hat{p}(x_i))\}_{i=1}^{N_{T}}$.
\end{algorithmic}
\end{algorithm}

\textbf{Self-Weighted Entropy Minimization.}\quad
After updating the memory, we replay the reliable samples stored in the memory $\mathcal{M}$ to optimize the model.
Following \cite{wang2020tent}, we employ entropy minimization as our loss function.
This self-supervised loss encourages the model to exhibit higher confidence in its predictions.
In order to give more valuable samples a higher importance in optimization, we weight samples in $\mathcal{M}$ in a self-weighted manner.
Given samples $\{x_i\}_{i=1}^{\lvert\mathcal{M}\rvert}$, we calculate the following loss function:
\begin{equation}
\label{equ:loss}
  % \begin{aligned}
    \mathcal{L}_{sem}(\mathcal{M}, \theta_t) =\sum\limits_{x_i\in \mathcal{M}} \frac{w_i}{\sum_j w_j}\cdot \mathcal{H}(f_{\theta_t}(x_i))
    % &= \sum\nolimits_i \sigma_i(-\mathcal{H}(p_1), \cdots,-\mathcal{H}(p_{\mathcal{M}})])\ \mathcal{H}(p_i), 
    % \sum\nolimits_i \ \mathcal{H}(p_i),
  % \end{aligned}
\end{equation}
where $w_i = e^{-\mathcal{H}(f_{\theta_t}(x_i))}$ denotes the weight of sample $x_i$.
% where $p_i = f_{\theta_t}(x_i)$, and $\sigma(\cdot)$ is the softmax function to normalize the weights.
\textbf{It's worth noting that $w_i$ is also involved with the gradient back-propagation.}
The reason for this adaptive weighting is two-fold: (i) it gives greater weights to samples with lower entropy, which are considered more reliable. (ii) The model needs to learn the weight of each sample and adaptively give important samples higher weights, instead of solely minimizing entropy for all samples. 

\textbf{Stable Optimization Strategy.}\quad To optimize the model in a stable way, we employ sharpness-aware minimization (SAM) \cite{foret2020sharpness} and step size decay in parameter updating.
SAM is a technique that aims to improve generalization performance in deep learning models by minimizing a loss function while simultaneously controlling the sharpness of the loss landscape.
Previous work \cite{foret2020sharpness} demonstrates that SAM exhibits resilience against noisy labels, which is also a prevalent challenge in TTA. 
Therefore, we consider SAM as a good choice of the optimizer in the adaptation scenario \cite{niu2022towards, gong2023sotta}.
Mathematically, this involves minimizing the following objective function:
\begin{equation}
\label{equ:sam}
  \min_{\theta_t}\mathcal{L}^{SA}_{sem}(\mathcal{M},\theta_t),\ \text{where}\ \mathcal{L}^{SA}_{sem}(\mathcal{M},\theta_t) \triangleq \max_{||\epsilon||_2\leq\rho}\mathcal{L}_{sem}(\mathcal{M}, \theta_t+\epsilon),
\end{equation}
where $\rho$ represents the radius of the neighborhood.
SAM minimizes the maximum loss within the parameters' vicinity.

Recent research \cite{press2024rdumb, lee2023towards} has revealed that under long-term adaptation, TTA algorithms may suffer from substantial error accumulation and consequent performance degradation.
This issue becomes even more severe when test data contains outliers.
To mitigate this problem, we introduce a test-time step size (\textit{a.k.a.} learning rate) decay strategy.
To achieve this, we update the model's parameters $\theta_t$ using cosine annealing decay step size:
\begin{equation}
\label{equ:ss}
  \theta_{t+1} = \theta_t - \alpha_t \nabla\mathcal{L}^{SA}_{sem}(\mathcal{M}, \theta_t),\ \text{where}\ \alpha_t = \frac{1}{2}\alpha_0(1+cos(\frac{t}{T}\pi)),
\end{equation}
where $\alpha_t$ is the step size at timestamp $t$ and $T$ is a hyper-parameter to control the step size decay. While step size decay is commonly utilized in model training \cite{you2019does}, we leverage it during testing for a distinct purpose.
In the early stages of TTA, the model should adapt swiftly due to the significant disparity between the source and target domains.
However, as adaptation progresses and the model acquires comprehensive distributional information about the target domain, it should reduce parameter updates to prevent error accumulation and enhance stability.
We summarize the pipeline of STAMP in \Cref{alg:STAMP}.

% \usepackage{multirow}
% \setlength{\tabcolsep}{1.0pt}
\begin{table*}[!t]
\centering
\small
\caption{Results (\%) on \textbf{CIFAR10-C} and \textbf{CIFAR100-C} datasets under outlier-aware TTA setting across \textbf{four outlier datasets}. We report the accuracy (ACC), AUC score, and H-score. All the results are averaged over 15 different corruptions. The results in bold denote the best performance.}
\label{table:cifar}
\resizebox{0.95\textwidth}{!}{
\begin{tabular}{llccccccccccccccc}
\toprule
                   &  & \multicolumn{3}{c}{Noise} & \multicolumn{3}{c}{SVHN-C} & \multicolumn{3}{c}{LSUN-C} & \multicolumn{3}{c}{TinyImageNet-C} &  \multicolumn{3}{c}{Avg.} \\ \cmidrule(r){3-5} \cmidrule(r){6-8} \cmidrule(r){9-11} \cmidrule(r){12-14} \cmidrule(r){15-17}                   &   &  ACC     & AUC      &H-score      &       ACC     & AUC      &H-score      &        ACC     & AUC      &H-score      &     ACC     & AUC      &H-score      &     ACC     & AUC      &H-score           \\ \midrule
\multirow{10}{*}{\rotatebox{90}{CIFAR10-C}} & Source  &  57.3     & 70.4      &  62.3    &  57.3     &   67.4    &  61.1    &  57.3     &  62.8     &  59.6   &57.3       & 64.5     &59.4       & 57.3  & 66.3   & 60.6   \\

                   & BN Stats\cite{nado2020evaluating}  &  72.9     & 68.6      & 70.6     &  78.7     &  75.3     & 76.9     & 79.4      & 79.4      &  79.4    &  79.0     & 72.9      & 75.8     &  77.5     &  74.0     & 75.7     \\
                   
                   & Tent\cite{wang2020tent} &  77.4     &  48.7     &  59.7    &     80.8  &   54.9    &   65.1   &    81.2   &     62.3  &   70.4   &   81.1    &  65.6     &    72.4  &  80.1     &   57.9    &  66.9    \\
                   
                   & EATA\cite{niu2022efficient} &   72.9    &  68.5     &70.6      &78.8       & 75.3      & 76.9     &  79.4     &  79.4     &  79.4    &   78.9    &   73.1    &  75.9    &    77.5   &    74.1   &    75.7  \\
                   
                   & SAR\cite{niu2022towards} &    72.9   &  68.5      &70.6      & 78.7       & 75.3      & 76.9     &  79.4     &  79.4    &79.4 & 79.0     & 72.9      &   75.8    &  77.5    &   74.0    &      75.7      \\
                   
                   & CoTTA\cite{wang2022continual} &  77.3     &  62.4     &  67.3    &    81.6   &    78.6   &  80.1    & 82.2      &     84.2  &  83.2    &   81.9    &   \textbf{75.3}    &  78.4    &  80.8     &  75.1     &  77.2    \\
                   
                   & RoTTA\cite{yuan2023robust}    &   77.6    &  74.3     &   75.6   &    78.4   &   76.0    &  77.2    &   78.8    &    79.5   & 79.1     &    78.6   &  73.3     &    75.8  &    78.3   &   75.8    &    76.9  \\
                   
                   & SoTTA\cite{gong2023sotta} &  77.8     &51.7       &61.6      &79.3       &72.8       &75.9      &79.8       & 77.9      & 78.8     & 79.6      & 72.6      & 75.9     &  79.1     & 68.8      & 73.1     \\
                   
                   & OWTTT\cite{li2023robustness} &  62.3     &  64.4     &  58.5    &    66.1   &   75.3    &  69.6    &    63.1   &        78.9    &   68.5    &    56.3   & 58.8     &  56.2     &  62.0     &  69.3  & 63.2  \\ \rowcolor{lightcyan}
                 \cellcolor{white}
                   &  STAMP &   \textbf{77.9}    &    \textbf{83.2}   &   \textbf{80.1}   &     \textbf{82.3}  &   \textbf{79.2}    &  \textbf{80.6}    &   \textbf{83.5}    &    \textbf{86.3}   &  \textbf{84.8}    &   \textbf{82.6}    &  74.9     &   \textbf{78.5}   &  \textbf{81.6}     &  \textbf{80.9}     &   \textbf{81.0}   \\ \midrule
\multirow{10}{*}{\rotatebox{90}{CIFAR100-C}} & Source  &  35.8     &   43.1    &  38.0    &    35.8   &  49.4     &  40.1    &  35.8     &    58.2   &   43.2   &    35.8   &  57.1     & 42.7     &    35.8   &   52.0    &   41.0   \\

                   & BN Stats\cite{nado2020evaluating} &  45.8     &    80.9   &   58.4   &  52.7     &   72.5    &   60.9   &  53.7     &   73.8    &  62.0    &    53.2   &   68.6    &   59.7   &  51.3     &    74.0   &    60.3  \\
                   
                   & Tent\cite{wang2020tent}&   47.9    &   55.8    &   51.2   &  54.4     &  70.4     &  61.2    &    55.4   &    72.4   & 62.7     &   55.0    &    68.6   & 60.9     &  53.2     &  66.8     &    59.0  \\
                   
                   & EATA\cite{niu2022efficient}  &    55.2   &  86.1     &   67.1   &     58.1  &   75.6    &  65.6    &  58.8    &    77.2   &   66.7   &    58.6   &   70.7    &    64.0  &    57.7   &   77.4    &    65.9  \\
                   
                   & SAR\cite{niu2022towards} &   57.5    &    88.6   &  68.9    &     59.2  &  65.2     &   61.9   &   60.5    &    73.5   &   66.3   &     60.8  &   72.1    & 65.9      &    59.5   &     74.9  &  65.8    \\
                   
                   & CoTTA\cite{wang2022continual} &   47.0    & 83.4      &  59.9    &     53.7  &   73.2    &    61.8  &  54.3     &    76.9   &  63.6    &  54.5     &   68.1    &    60.4  &  52.4     &  75.4     &  61.4    \\
                   
                   & RoTTA\cite{yuan2023robust} &   47.9    &   54.0    &  49.4    &     47.3  &   67.0    &   55.3   &  48.3     &     69.5  &   56.7   &    47.8   &    65.5   &    55.0  &   47.8    &  64.0     &  54.1    \\
                   
                   & SoTTA\cite{gong2023sotta} &   54.4    &  53.3     & 52.8     &     53.6  &    70.3   &  60.7    &   54.4    &     70.8  &   61.4   &     53.9  &    68.4   &   60.1   &     54.1  &    65.7   &   58.8   \\
                   
                   & OWTTT\cite{li2023robustness} &  47.1     &  70.3     &   56.2   &      53.9 &    74.3   &   62.3   &   54.5    &      73.5 & 62.5     &   54.2    &    68.5   &   60.4   &    52.4   &    71.6   &   60.4   \\
                   \rowcolor{lightcyan}
                 \cellcolor{white}  &  STAMP&   \textbf{57.9}    & \textbf{98.4}      & \textbf{72.8}     &     \textbf{63.7}  &   \textbf{82.1}    &   \textbf{71.7}   &  \textbf{63.7}     &     \textbf{82.6}  &  \textbf{71.9}    &   \textbf{63.9}    &   \textbf{75.5}    &    \textbf{69.2}  &   \textbf{62.3}    &   \textbf{84.7}    &  \textbf{71.3}   \\ \bottomrule

\end{tabular}
}
\end{table*}
\section{Experiments}
\subsection{Setup}
\textbf{Datasets.}\quad
Following previous TTA methods \cite{wang2022continual, gong2022note, goyal2022test}, we adopt three corruption datasets as normal datasets (\ie, in-distribution datasets).
CIFAR10-C and CIFAR100-C \cite{krizhevsky2009learning, hendrycks2019benchmarking} stand out as the most widely used benchmarks in TTA tasks, comprising 10/100 categories with 15 types of corruptions respectively.
Each type of corruption has 10,000 images.
Similarly, ImageNet-C \cite{deng2009imagenet, hendrycks2019benchmarking} serves as a significant benchmark, consisting of 1,000 categories with the same 15 types of corruption.
% It is constructed by applying the corruptions same as CIFAR10-C to the validation set of ImageNet \cite{deng2009imagenet}.
For out-of-distribution datasets, we choose Noise, SVHN \cite{netzer2011reading}, LSUN \cite{yu2015lsun}, and TinyImageNet \cite{le2015tiny} for the CIFAR10/100-C datasets, and Textures \cite{cimpoi2014describing} and Places365 \cite{zhou2017places} for ImageNet-C.
To maintain consistency in domain shift between outliers and normal datasets, we apply the same corruptions~\footnote{\url{https://github.com/hendrycks/robustness}} on outlier datasets.
(\eg, SVHN-C refers to the corruptions version of SVHN dataset.)
Throughout all experiments, we maintain a fixed ratio of 8:2 between normal samples and outliers.
Further details of the outlier datasets are provided \Cref{sec:setup}.

\setlength{\tabcolsep}{7.0pt}
\begin{table}[!t]
\centering
\small
\caption{Results (\%) on \textbf{ImageNet-C} under outlier-aware TTA setting across \textbf{two outlier datasets}. All the results are averaged over 15 different corruptions.}
\label{table:imagenet}
\resizebox{0.65\linewidth}{!}{
\begin{tabular}{lcccccc}
\toprule
& \multicolumn{3}{c}{Places365-C} & \multicolumn{3}{c}{Textures-C} \\ \cmidrule(r){2-4} \cmidrule(r){5-7}
  Method&     ACC  &  AUC      & H-score     & ACC    & AUC      & H-score    \\ \midrule
 Source      &   18.2    &  61.6    &  26.1     &  18.2     &  54.6 &25.8    \\
 BN Stats\cite{nado2020evaluating}&  31.1     &67.7       &41.1      &  31.6     & 61.2      &   40.7   \\
 Tent\cite{wang2020tent}&    34.9   &  51.8     & 39.5     &  39     & 48.6      &  42.0    \\
 EATA\cite{niu2022efficient}&  \textbf{46.4}     &   72.6    &   56.0   & 46.4      & 62.2      & 52.8     \\
 SAR\cite{niu2022towards}&   44.9    & 73.3      & 55.0     &   45.6    &  67.0     &   54.0   \\
 CoTTA\cite{wang2022continual}& 33.8      & 66.9      &    43.5  & 34.2      & 60.7      & 42.8     \\
 RoTTA\cite{yuan2023robust}&  36.6     & 68.6      & 46.5     &   37.0    &  65.3     &  46.5    \\
 SoTTA\cite{gong2023sotta}&   41.7    &  67.8     &50.7      &41.8       &60.3       &  48.8    \\
 OWTTT\cite{li2023robustness}&   9.1    & 54.0      &  13.9    &   9.4    &  59.4     &   14.6   \\
 \rowcolor{lightcyan}
STAMP &   \textbf{46.4}   &\textbf{77.7}       & \textbf{57.6}     & \textbf{46.5}      &   \textbf{71.9}    & \textbf{56.2}\\ \bottomrule   
\end{tabular}
}
\end{table}

\textbf{Baselines.}\quad
To have a comprehensive comparison of existing TTA methods, we select eight TTA algorithms as our baselines.
\textbf{Source} refers to the source pre-trained model.
\textbf{BN Stats} \cite{nado2020evaluating} adjusts batch normalization (BN) layers \cite{ioffe2015batch} with the statistics of test data.
\textbf{Tent} \cite{wang2020tent} optimizes the BN affine parameters in the model by entropy minimization.
\textbf{CoTTA} \cite{wang2022continual} calculates weighted-average and augmentation-averaged pseudo-label to distill knowledge from the teacher model to the student model, and stochastically restores parameters to prevent the model from catastrophic forgetting.
\textbf{EATA} \cite{niu2022efficient} adopts the entropy and diversity weight with elastic weight consolidation \cite{kirkpatrick2017overcoming} for anti-forgetting.
\textbf{SAR} \cite{niu2022towards} excludes high-entropy samples from sharpness-aware entropy minimization and reset the model when it overfits.
\textbf{RoTTA} \cite{yuan2023robust} uses robust batch normalization and category-balanced sampling with timeliness and uncertainty to deal with the correlated label problem.
\textbf{SoTTA} \cite{gong2023sotta} applies high-confidence uniform-class sampling and entropy-sharpness minimization to alleviate the impact of noisy samples.
\textbf{OWTTT} \cite{li2023robustness} proposes prototype expansion with prototype clustering and distribution alignment. 
% In addition, it is equipped with an adaptive threshold OOD detector based on prototypes to detect outliers.

\textbf{Evaluation Metrics.}\quad
We evaluate the above baselines for both generalization and outlier detection.
For generalization, we adopt classification accuracy (ACC) as a metric.
For outlier detection, we consider the area under the ROC curve (AUC) \cite{davis2006relationship}, a threshold-free metric to measure the binary classification problem.
The ROC curve depicts the relationship between the true positive rate (TPR) and the false positive rate (FPR).
Models with better detection ability will have a higher AUC score.
To consider both generalization and detection performance simultaneously, we also report H-score:
\begin{equation}
  H\text{-}score = \frac{2\cdot ACC\cdot {AUC}}{ACC+{AUC}}
\end{equation}

\textbf{Implementation Details.}\quad
We adopt the corruption with severity 5 in all experiments.
We consider ResNet-18 \cite{he2016deep} as backbones for CIFAR10-C, CIFAR100-C and ResNet-50 \cite{he2016deep} for ImageNet-C.
For CIFAR10-C and CIFAR100-C, we train the source model using standard cross-entropy loss with the initial learning rate of $0.1$ and multi-step learning rate scheduling.
For ImageNet-C, we use the released checkpoint from TorchVision library \cite{torchvision2016torchvision}.
For a fair comparison, we set the batch size for all methods as $64$ and the capacity of our memory bank is $64$.
Concerning the hyperparameters, we search the hyperparameters on a validation set and set the learning rate to $0.1/0.05/0.01$, $\mathcal{H}_{thr}$ to $0.25/0.9/0.8 \times \ln\lvert\mathcal{C}_S\rvert$, $T$ to $150/150/750$ for CIFAR10-C/CIFAR100-C/ImageNet-C.
Moreover, we set the other hyperparameter the same in experiments with $\beta = 0.1, \rho = 0.05, K=16$. 
For parameters' updating, we select the affine parameters in the BN layer for optimization and freeze the top layer’s parameter following \cite{niu2022towards}.
More details of hyper-parameters searching and setting of baselines and STAMP can be found in \Cref{sec:setup}.

\setlength{\tabcolsep}{7.0pt}
\begin{table}[!t]
\centering
\small
\caption{Results (\%) on \textbf{CIFAR10/100-C} datasets under outlier-aware TTA setting.
We divide datasets into the normal part and the outlier part. In detail, we choose $8/80$ classes as known samples and other $2/20$ classes as outliers for CIFAR10/100-C. All the results are averaged over 15 different corruptions.}
\label{table:open}
\resizebox{0.6\linewidth}{!}{
\begin{tabular}{lcccccc}
\toprule
& \multicolumn{3}{c}{CIFAR10-C (8:2)} & \multicolumn{3}{c}{CIFAR100-C (80:20)} \\ \cmidrule(r){2-4} \cmidrule(r){5-7}
  Method&     ACC  &  AUC      & H-score     & ACC    & AUC      & H-score    \\ \midrule
 Source      & 60.6    & 61.5    & 60.0     & 37.1     & 60.6 & 44.2   \\
 BN Stats\cite{nado2020evaluating}&  79.4    & 66.9     &  72.6   &  55.5  & 66.2     &    60.2 \\
 Tent\cite{wang2020tent}&   80.1  &   65.9   &  72.3  & 60.4      & 65.6      & 62.8    \\
 EATA\cite{niu2022efficient}&   81.5   &67.9     &  74.1   &   61.3   &  67.4    &  64.2   \\
 SAR\cite{niu2022towards}&  82.3   &  66.0   &    73.2 &   63.4  &  69.1   & 66.1   \\
 CoTTA\cite{wang2022continual}&    81.8  &  65.4   & 72.5   & 57.1     &  66.3    &  61.3    \\
 RoTTA\cite{yuan2023robust}& 79.4    & 62.4    & 69.8   & 50.6    &63.7      &56.2     \\
 SoTTA\cite{gong2023sotta}& 82.7    & 62.7    &  71.3  &  61.4  &  68.2   &  64.6  \\
 OWTTT\cite{li2023robustness}&   65.9   & 63.5  & 64.4  & 56.2   & 66.0    & 60.6  \\
\rowcolor{lightcyan} STAMP &    \textbf{85.0}  &   \textbf{69.4}  &   \textbf{76.4}  & \textbf{66.0}    &\textbf{71.2}     &\textbf{68.5}  \\ \bottomrule   
\end{tabular}
}
\end{table}

\subsection{Results}
\textbf{Results under outlier-aware TTA setting.}\quad 
We first evaluate TTA methods on the datasets with outliers.
The results for CIFAR datasets are illustrated in \Cref{table:cifar}.
We observe that a simple baseline method, Tent \cite{wang2020tent}, has a competitive generalization performance on distribution shift (achieving an accuracy of $80.1\%$ on CIFAR10-C).
However, it exhibits substantial degradation in the AUC score compared to Source, for instance, dropping from $70.4\%$ to $48.7\%$ on CIFAR10-C with Noise outlier dataset.
This decline could be attributed to Tent's indiscriminate minimization of entropy across all samples, leading to a gradual loss of its ability to discern outliers.
In addition, our experimental findings demonstrate that with judicious parameter selection, TTA methods can enhance not only the generalization capability against distribution shift but also the detection ability for outliers.
Compared to existing TTA methods, STAMP achieves the most favorable performance on both classification and outlier detection.
On CIFAR10-C, STAMP achieves the highest ACC, and $+5.1\%$ huge improvements in AUC compared to the previous methods.
Regarding CIFAR100-C, STAMP exhibits $+2.8\%$ and $+7.3\%$ improvements compared to the most effective TTA methods on ACC and AUC, respectively. We also report the results on the ImageNet-C in \Cref{table:imagenet}. STAMP still achieves the highest performance. Specifically, we outperform existing TTA methods with $+1.6\%, +2.2\%$ on H-score with Places365-C and Textures-C outlier datasets, respectively.

Additionally, we evaluate the TTA algorithm under a more challenging scenario involving homogeneous outlier datasets in \Cref{table:open}, which makes it more difficult to identify outliers.
STAMP also exhibits the most promising performance on CIFAR10-C and CIFAR100-C under this scenario.

Concerning the above results, we achieve optimal results in most cases, affirming the robustness and effectiveness of STAMP.

\setlength{\tabcolsep}{5.0pt}
\begin{table*}[!t]
\centering
\small
\caption{Classification accuracies (\%) on \textbf{CIFAR10-C} and \textbf{CIFAR100-C} under closed-set TTA setting.}
\label{table:clean}
\resizebox{0.95\textwidth}{!}{
\begin{tabular}{llm{0.5cm}m{0.5cm}m{0.5cm}m{0.5cm}m{0.5cm}m{0.5cm}m{0.5cm}m{0.5cm}m{0.5cm}m{0.5cm}m{0.5cm}m{0.5cm}m{0.5cm}m{0.5cm}m{0.5cm}c}
\toprule
\multicolumn{2}{c}{Method} &\rotatebox{70}{Gaussian}  &\rotatebox{70}{shot}  & \rotatebox{70}{impulse} &\rotatebox{70}{defocus}  & \rotatebox{70}{glass} & \rotatebox{70}{rotion} & \rotatebox{70}{zoom} &\rotatebox{70}{snow}  & \rotatebox{70}{frost} & \rotatebox{70}{fog} & \rotatebox{70}{brightness} & \rotatebox{70}{contrast} & \rotatebox{70}{elastic} &\rotatebox{70}{pixelate}  &\rotatebox{70}{jpeg}  & Avg.  \\ \midrule
\multirow{10}{*}{\rotatebox{90}{CIFAR10-C}} & Source & 28.7	&35.2	&24.2	& 57.3&49.0	&66.4	&64.8	&76.7	&62.9	&73.4	&	90.3&	31.4&	78.8&	46.5&74.7 & 57.3
 \\ 
& BN Stats\cite{nado2020evaluating}  & 70.2	& 72.0&63.6	&87.6	&66.5	&85.9 &86.9	&82.1	&80.5	&84.1	&90.8	&85.4	&77.3 &78.6	&74.3	&79.0
 \\
&Tent\cite{wang2020tent} &	76.1& 78.4	&70.5	&87.8	&70.1	&86.8	&87.5	&84.8	&82.0	&85.1	&91.1	&86.6	&79.4&	82.7&78.5 &81.8
 \\
&CoTTA\cite{wang2022continual} & 76.1&77.6	&73.9	&87.9	&71.9&	86.9&87.5	&82.8	&82.0	&85.2	&90.8	&85.7	&79.3	&80.4	&78.3	&81.8
  \\
& EATA\cite{niu2022efficient} & 76.5	&77.9	&71.7	&89.1	&70.4 &87.4	&\textbf{89.0}	&85.2	&84.2	&86.0	&91.5	&\textbf{88.4} &79.9	&83.9	&78.5	&82.6
  \\
& SAR\cite{niu2022towards} &70.7	&72.1	&66.8	&87.6	&68.2	&85.9	&86.9	&82.1	&80.5	&84.1	&90.8	&85.4	&77.3&	78.6&74.3 &79.4
 \\
 &RoTTA\cite{yuan2023robust} & 69.6	&71.0	&62.9	&87.5&	67.0&	85.9&86.9	&82.4	&79.6	&84.7	&91.2	&73.5	&78.3&77.9	&74.9	&78.2
 \\
& SoTTA\cite{gong2023sotta} &75.8	&79.2	&71.5	&\textbf{89.4}	&70.6	&\textbf{87.8}	&\textbf{89.0}	&85.6	&84.0	&87.2	&\textbf{92.4}	&87.4	&79.9	&84.5	&79.0	&82.9
  \\

& OWTTT\cite{li2023robustness}  &70.1	&72.0	&63.0	&86.9	&66.5	&85.7	&86.7	&82.7&	80.9&	84.5&91.2	&83.0	&77.9	&76.8	&74.9	&78.9
 \\ \rowcolor{lightcyan} \cellcolor{white}
& STAMP &\textbf{80.9}	&\textbf{82.9}	&\textbf{77.2}	&87.6	&\textbf{74.9}&	86.6&87.7 &\textbf{85.9}&\textbf{85.9}&\textbf{88.1}&90.4&87.2&\textbf{80.3}&\textbf{86.7}&\textbf{82.6}&\textbf{84.3}
 \\ \midrule
\multirow{10}{*}{\rotatebox{90}{CIFAR100-C}} & Source &12.4	&14.5	&7.2	&36.0 &44.7	&45.1	&45.2	&49.5	&41.6	&36.9	&63.3	&13.2	&57.5	&23.8	&46.6	&35.8
 \\ 
& BN Stats\cite{nado2020evaluating}  &41.1	&41.3 &38.9	&63.1	&51.5	&60.8 &63.8	&51.2	&53.5	&53.2	&64.4	&59.0	&58.4 &58.0&47.6	&53.7
 \\
&Tent\cite{wang2020tent} &52.3	&52.1	&47.7	&66.9	&56.1	&64.3	&65.3	&58.3	&58.7	&60.0	&67.8	&62.1	&61.8&	63.0   &54.5 &59.4
 \\
&CoTTA\cite{wang2022continual} & 47.0&47.8	&45.1	&59.6	&54.0&59.5	&61.3	&53.3	&55.0	&52.9	&62.6	&50.3	&58.1	&61.8	&53.1	&54.8
  \\
& EATA\cite{niu2022efficient} &50.7	&53.5	&48.1	&67.0	&55.6 &64.8	&67.0	&59.1	&59.2	&60.4	&67.7	&63.9 &61.8	&63.3	&54.6	&59.8
  \\
& SAR\cite{niu2022towards} &\textbf{55.1}	&55.0	&51.2	&68.4	&58.2	&66.0	&67.4	&60.3	&60.8	&61.9	&69.8	&65.5	&63.6&66.2	&56.8 &61.7
 \\
& RoTTA\cite{yuan2023robust} &35.9	&36.6	&33.8	&60.6	&47.1	&57.7	&60.8	&48.0	&42.2	&50.8	&59.2	&32.1	&53.8	&52.3	&44.4	&47.7
  \\
&SoTTA\cite{gong2023sotta} &51.6	&53.8	&47.4	&66.9&	56.9&	65.3&	68.1&	58.9&	60.1&	60.1&	69.4&	63.1&62.3&62.8	&54.8	&60.1
 \\
& OWTTT\cite{li2023robustness}  &41.6	&42.8	&38.8	&63.4	&52.6	&61.6	&64.8	&53.3&	54.8&	54.5&	65.8&	58.4&	60.1&57.6	&49.3	&54.6
 \\ \rowcolor{lightcyan} \cellcolor{white}
& STAMP &\textbf{57.2}	&\textbf{58.5}	&\textbf{52.8}	&\textbf{69.9}	&\textbf{61.4}&	\textbf{68.1}& \textbf{70.1}&\textbf{63.3}&\textbf{63.9}&\textbf{64.8}&\textbf{72.2}&\textbf{69.9}&\textbf{66.5}&\textbf{69.2}&\textbf{59.0}&\textbf{64.4}
 \\ 
 \bottomrule
\end{tabular}
}
\end{table*}

\textbf{Results under closed-set TTA setting.}\quad
Additionally, we assess our method in a scenario without outliers as most TTA works do.
We present the classification accuracy on the clean dataset in \Cref{table:clean}.
Remarkably, our method attains the highest accuracy on both CIFAR10-C and CIFAR100-C datasets, achieving an average improvement of $+2.9\%$ over the second-best method (SoTTA).
This shows that STAMP still has strong applicability even in the traditional situation without outliers.

To summarize, all the above results prove that our method achieves the best results under both outlier-aware TTA and conventional closed-set settings.

\begin{figure}[!b]
\begin{center}
\centerline{\includegraphics[width=0.5\linewidth]{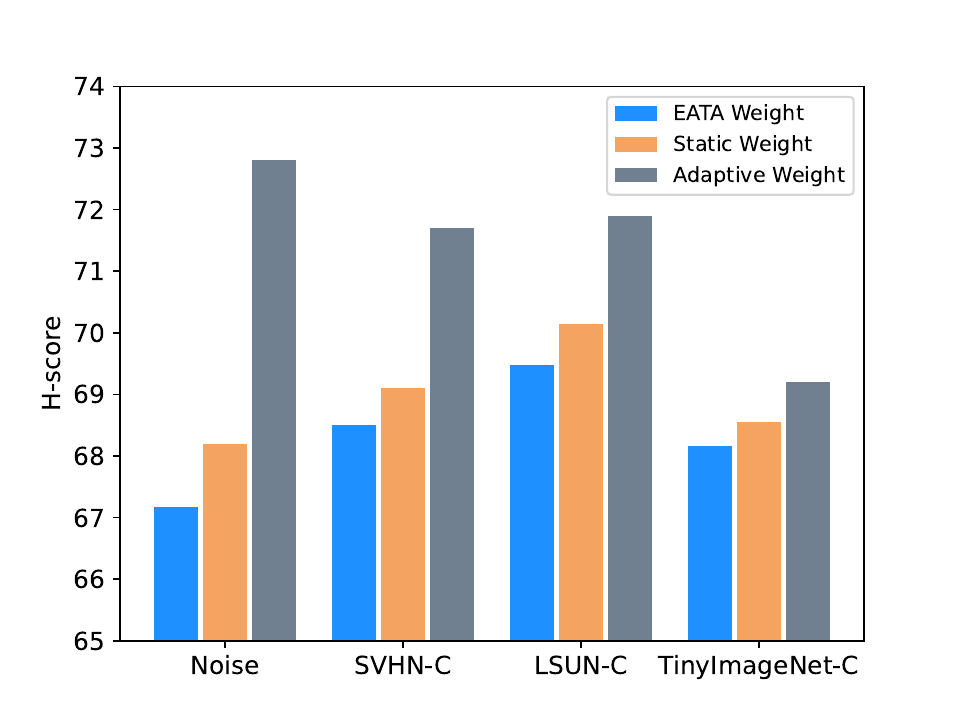}}
\caption{Results for the different weight strategies on dataset CIFAR100-C with SVHN-C outlier dataset. Static weight represents that the $w_i$ in \Cref{equ:loss} without gradient back-propagation. EATA weight represents the weight for reliable samples used in EATA \cite{niu2022efficient}.}
\label{fig:weight}
\end{center}
\vskip -0.2in
\end{figure}

\tabcolsep=0.3cm
\begin{table}[tbp]
\centering
\small
\caption{Ablation studies on reliable class-balanced memory (RBM), self-weighted entropy minimization (S.W.), sharpness-aware minimization (S.A.), and decay step size (D.S.). Experiments are conducted on \textbf{CIFAR100-C} with the outlier dataset \textbf{SVHN-C}. The first row in the table reports the results of entropy minimization.}
\label{table:ablation}
\resizebox{0.5\linewidth}{!}{
\begin{tabular}{cc|cc|cccc}
\toprule
 S.A.   & D.S.  & RBM & S.W. & ACC  &AUC  &H-score  \\ \midrule
 -  &-  &-&-&54.4  &70.4  &61.2  \\
 -  &\checkmark  &-&-&58.9  &65.3  & 61.8 \\
 \checkmark  &-  &-&-&56.4  &72.8  & 63.5 \\
\checkmark   &\checkmark  &- &-&59.2  &70.8  & 64.4 \\ 
% \midrule
  \checkmark   &\checkmark&-&\checkmark  &59.3  &77.7  & 67.1 \\
 \checkmark   &\checkmark  &\checkmark &-&61.5  & 75.1 &67.6  \\
  \checkmark  &\checkmark  &\checkmark&\checkmark&63.7  &82.1  &71.7 \\ \bottomrule

\end{tabular}
}
\end{table}

\subsection{Ablation Study}
\label{sec:ablation}
\textbf{Effect of each component.}\quad We delve into the role of each component within our method as follows.
First, we provide the results of different weight strategies of entropy minimization in \Cref{fig:weight}.
We compare our adaptive weight with static weight and EATA weight \cite{niu2022efficient}. EATA calculates the weight by $exp[\mathcal{H}_{thr}-\mathcal{H}(p(x))]$ and static weight calculates weight in the same way of \Cref{equ:loss}. Different from adaptive weight, both EATA and static weight don't involve any gradient back-propagation.
As depicted in \Cref{fig:weight}, adaptive weight shows the best performance among the strategies.

Furthermore, we conduct an ablation study on each component of our method, as summarized in \Cref{table:ablation}.
We observe that the effect of combining test-time step size decay and SAM is better than using either of them alone.
Step size decay can improve the ACC but result in AUC decline.
However, combining SAM and step size decay mitigates the decline in AUC.
In addition, after the introduction of self-weighted entropy minimization, the AUC of the model further increased to $77.7\%$.
Notably, by removing reliable class-balanced memory from our method, the ACC decreases from $63.7\%$ to $59.3\%$.
It reveals that reliable class-balanced memory significantly contributes to enhancing the generalization ability to distribution shift.
Moreover, by removing the self-weighted entropy minimization in STAMP, its AUC drops from $82.1\%$ to $75.1\%$.
Self-weighted entropy minimization plays an important role in outlier detection.
Besides decreasing the sample entropy, self-weighted entropy minimization incorporates entropy into the optimization weight, allowing outliers that deviate from the majority of samples to receive lower weights, consequently increasing their entropy. This mechanism enhances outlier detection performance.
\tabcolsep=0.3cm
\begin{table}[!t]
    \centering
    \small
    \caption{H-score (\%) with the different ratios of outliers to total samples on \textbf{CIFAR100-C} with different outlier datasets.}
    \label{table:ratio}
    \resizebox{0.5\linewidth}{!}{
\begin{tabular}{cccc}
\toprule
 & SVHN-C & LSUN-C & TinyImageNet-C \\
 \midrule
 % Source& 40.1  & 43.2    & 42.7     \\ 
 % \midrule
  5\%&   69.2 & 70.8  & 68.7     \\
 10\%&  70.9     &  71.5    & 69.2  \\
 20\%&  71.7     &  71.9     &  69.2    \\
 33\%&  71.0     &  70.7     &  68.5      \\
 50\%&  70.0     &  69.5     &  67.8     \\ 
 \bottomrule
\end{tabular}
}
\end{table}

\textbf{Effect of outlier proportion.}\quad
In the realistic scenario, the proportion of outliers can vary considerably.
We investigate how this variability affects the performance of our method.
Specifically, we manipulate the proportion of outliers, ranging from $5\%$ to $50\%$, and present the corresponding H-scores in \Cref{table:ratio}.
The results show that the fluctuation of our method is within $3\%$.
The findings suggest that STAMP exhibits robustness across different scenarios with varying proportions of outliers, as it does not experience significant degradation.

\section{Conclusion}
In this study, we research the challenge of handling outliers in test-time adaptation.
To tackle this issue effectively, we introduce a scenario called outlier-aware test-time adaptation, which refers to conducting both recognition and outlier rejection simultaneously.
Additionally, we propose STAMP, using augmentation-averaged estimation and a filtering mechanism to store reliable samples in memory.
Moreover, we dynamically discard samples with the highest class frequency from memory to maintain a class-balanced prediction.
Then we replay the samples stored in the memory with self-weighted entropy minimization.
Furthermore, we leverage sharpness-aware minimization and learning rate decay to bolster the stability of our algorithm.
Through extensive experiments across multiple benchmarks, we demonstrate that our method surpasses existing TTA methods, underscoring its effectiveness and versatility in open-world applications.
\newpage
\section*{Acknowledgements}
% to be modified
We thank Yuhe Ding and Zhengbo Wang for their valuable feedback on our work.
This work was funded by the Beijing Municipal Science and Technology Project (No. Z231100010323005), the Beijing Nova Program (No. Z211100002121108), the Young Elite Scientists Sponsorship Program by CAST (2023QNRC001), and the National Natural Science Foundation of China under (No. 62276256).

\bibliographystyle{splncs04}
\bibliography{main}

\newpage
\appendix
% \onecolumn
\section{More Details of Experimental Setup}
\label{sec:setup}
\subsection{Outlier Datasets}
We introduce the details of the outlier datasets we used in this section.
\textbf{SVHN} \cite{netzer2011reading} dataset is a digital collection that includes 50000 images for training and 10000 images for testing.
\textbf{LSUN} \cite{yu2015lsun}
is a collection of large-scale labeled images designed for various scene understanding. In our experiment, we resize the image to $32\times 32$.
\textbf{TinyImageNet} \cite{le2015tiny} is an image classification dataset has 200 classes. Each class has 500 training images, 50 validation images, and 50 testing images.
\textbf{Textures} \cite{cimpoi2014describing} is composed of 5,640 texture images that are classified into 47 categories.
\textbf{Places365} \cite{zhou2017places} is a scene recognition dataset containing over 10 million images in total, including over 400 unique scene categories.

\begin{sloppypar}
\subsection{Additional Implement Details}
\textbf{Hyperparameters}.\quad
For all baseline methods and our method, we tune the hyperparameter according to the H-score on the Gaussian Noise corruption with Noise outliers.
The tuned hyperparameters are applied to all the domains and outlier datasets.
We list the tuned hyperparameters for each method as follows:

\textbf{Ours}\quad
For augmentation, we apply the same augmentation for CIFAR10-C following \cite{wang2022continual}.
For other datasets, we apply random crop and random horizontal flip augmentation.
In addition, since there are 1,000 classes in ImageNet, requiring the source model to be consistent with the current model output will result in filtering out a large number of samples, so we removed consistency filtering on the ImageNet experiment.

\textbf{Tent}\quad For Tent \cite{wang2020tent}, we set the learning rate $LR = 0.001/0.0001/0.00025$ for CIFAR10-C/CIFAR100-C/ImageNet-C. We referred to the official code~\footnote{https://github.com/DequanWang/tent} for implementations.

\textbf{CoTTA}\quad For CoTTA \cite{wang2022continual}, we set the learning rate $LR = 0.001/0.001/0.0001$, the restore constant $p = 0.005/0.005/0.00005$ and the threshold $p_{th}=0.95/0.9/0.05$ for CIFAR10-C/CIFAR100-C/ImageNet-C. We referred to the official code~\footnote{https://qin.ee/cotta} for implementations.

\textbf{EATA}\quad For EATA \cite{niu2022efficient}, we set the learning rate $LR = 0.001/0.001/0.0005$, the cosine sample similarity threshold $\epsilon = 0.2/0.1/0.025$, the entropy threshold $E_0 = 0.4\times\ln\lvert \mathcal{C}_S \rvert $, and the $Fisher_\alpha = 1$ for CIFAR10-C/CIFAR100-C/ImageNet-C. We referred to the official code~\footnote{https://github.com/mr-eggplant/EATA} for implementations.

\textbf{SAR}\quad For SAR \cite{niu2022towards}, we set the learning rate $LR = 0.005/0.005/0.0005$, the reset threshold $\tilde\Theta = 0.3/0.2/0.1$ and the entropy threshold $E_0 = 0.4\times\ln\lvert \mathcal{C}_S \rvert $ for CIFAR10-C/CIFAR100-C/ImageNet-C. We referred to the official code~\footnote{https://github.com/mr-eggplant/SAR} for implementations.

\textbf{RoTTA}\quad For RoTTA \cite{yuan2023robust}, we set the learning rate $LR = 0.005$ and keep other hyperparameter same as the original paper. We referred to the official code~\footnote{https://github.com/BIT-DA/RoTTA} for implementations.

\textbf{SoTTA}\quad For SoTTA \cite{gong2023sotta}, we set the learning rate $LR = 0.0001$, the confident threshold $C_0 = 0.9/0.5/0.33$ for CIFAR10-C/CIFAR100-C/ImageNet-C and keep other hyperparameter same as the original paper. We referred to the official code~\footnote{https://github.com/taeckyung/SoTTA} for implementations.

\textbf{OWTTT}\quad For OWTTT \cite{li2023robustness}, we set the learning rate $LR = 0.001/0.00001/0.000025$ and the domain alignment coefficient $\lambda = 5/1/1$ for CIFAR10-C/CIFAR100-C/ImageNet-C and keep other hyperparameter same as the original paper. We referred to the official code~\footnote{https://github.com/Yushu-Li/OWTTT} for implementations. In addition, we directly adopt the score proposed by OWTTT $os_i$ as the OOD score.
\end{sloppypar}
\begin{figure}[htbp]
  \centering
  \includegraphics[width=0.7\linewidth]{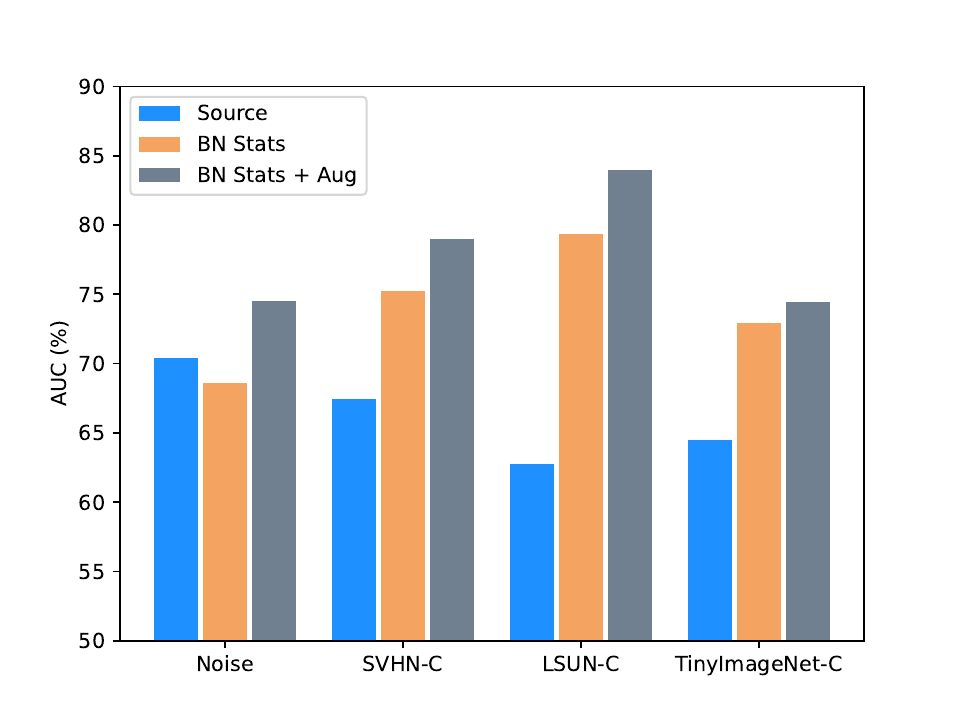}
  \caption{The illustration of the effectiveness of augmentation-averaged robust estimate on \textbf{CIFAR10-C}. $+ Aug$ means augmenting the data 16 times and averaging their results.}
  \label{fig:aug}
\end{figure} 
\section{Additional Results}
\subsection{Discussion about Augmentation-averaged Robust Estimation}
\label{sec:aug}
We showcase how augmentation-averaged robust estimation enhances the detection of outliers in \Cref{fig:aug}.
As depicted in the figure, by subjecting images to multiple augmentations and subsequently averaging their outputs, the impact of uncertainty within the sample is mitigated, thereby increasing the performance of outlier detection.

\end{document}